\documentclass{article}

\PassOptionsToPackage{numbers, compress}{natbib}

\usepackage[preprint]{neurips_2024}

\usepackage[utf8]{inputenc} 
\usepackage[T1]{fontenc}    
\usepackage{hyperref}       
\usepackage{url}            
\usepackage{booktabs}       
\usepackage{amsfonts}       
\usepackage{nicefrac}       
\usepackage{microtype}      
\usepackage{xcolor}         
\usepackage{amsmath}
\usepackage{graphicx}
\usepackage{tikz}
\usepackage{subcaption}
\usepackage{algorithm}
\usepackage{algorithmic}
\renewcommand{\algorithmicrequire}{\textbf{Input:}}
\renewcommand{\algorithmicensure}{\textbf{Output:}}

\DeclareMathOperator*{\argmin}{arg\,min}

\newcommand{\method}{DFedDGM}

\newcommand{\myhline}{\noalign{\global\arrayrulewidth0.3mm}\hline
                      \noalign{\global\arrayrulewidth0.3pt \vspace{0.5mm} }}
\usepackage{makecell}

\title{Data-Free Federated Class Incremental Learning with Diffusion-Based Generative Memory}

\author{%
  Naibo Wang \\
  Institute of Data Science\\
  National University of Singapore\\
  Singapore \\
  \texttt{naibowang@comp.nus.edu.sg} \\
  \And
  Yuchen Deng \\
  Institute of Data Science\\
  National University of Singapore \\
  Singapore \\
  \texttt{dengyuchen.cc@gmail.com} \\
  \And
  Wenjie Feng \\
  Institute of Data Science\\
  National University of Singapore \\
  Singapore \\
  \texttt{wenjie.feng@nus.edu.sg} \\
  \And
  Jianwei Yin \\
  College of Computer Science and Technology \\
  Zhejiang University\\
  Hangzhou, China \\
  \texttt{zjuyjw@cs.zju.edu.cn} \\
  \And
  See-Kiong Ng \\
  Institute of Data Science\\
  National University of Singapore \\
  Singapore \\
  \texttt{seekiong@nus.edu.sg} \\
}

\begin{document}

\maketitle

\begin{abstract}

Federated Class Incremental Learning (FCIL) is a critical yet largely underexplored issue that deals with the dynamic incorporation of new classes within federated learning (FL). Existing methods often employ generative adversarial networks (GANs) to produce synthetic images to address privacy concerns in FL. However, GANs exhibit inherent instability and high sensitivity, compromising the effectiveness of these methods. In this paper, we introduce a novel data-free federated class incremental learning framework with diffusion-based generative memory (\method) to mitigate catastrophic forgetting by generating stable, high-quality images through diffusion models. We design a new balanced sampler to help train the diffusion models to alleviate the common non-IID problem in FL, and introduce an entropy-based sample filtering technique from an information theory perspective to enhance the quality of generative samples. Finally, we integrate knowledge distillation with a feature-based regularization term for better knowledge transfer. Our framework does not incur additional communication costs compared to the baseline FedAvg method. Extensive experiments across multiple datasets demonstrate that our method significantly outperforms existing baselines, e.g., over a 4\% improvement in average accuracy on the Tiny-ImageNet dataset.
\end{abstract}

\section{Introduction}


Federated learning (FL)~\citep{li2020federated, duan2023towards} is a promising paradigm that facilitates collaborative machine learning across multiple clients, allowing them to construct a unified global model without sharing their local datasets~\citep{wang2022collaborative, balkus2022survey}. FL notably enhances data privacy~\citep{gao2021privacy} and security~\citep{ma2020safeguarding}, while enabling the derivation of a model with superior inferential capacities compared to those models developed through individual client-based training ~\citep{yang2019federated}. FL has attracted significant attention in both research and industry communities such as healthcare~\citep{chen2023metafed}, autonomous driving~\citep{elbir2022federated}, and finance~\citep{long2020federated}.


Despite its wide application, most FL frameworks~\citep{li2020federatedprox, gao2022feddc, li2021fedbn} operate under assumptions that are too constraining for realistic scenarios. A prevalent assumption is that the local data distributions of clients are static and unchanging over time, which is rarely the case in real-world settings where data is often dynamic and changes with the environment~\citep{zhang2020class, zhao2020maintaining, mittal2021essentials}. For instance, in healthcare, models initially trained on historical disease data must generalize to newly emerging diseases. A relevant example is the necessity for models to develop in response to the new variants of COVID-19~\citep{ciotti2020covid, zhang2021dynamic}, which continue to evolve because of the virus's high mutation rate. Therefore, it is crucial for models to quickly adapt to new data while maintaining performance on previous data distributions.


An intuitive solution to address the challenge of continuously emerging data classes is to train new models from scratch. However, this approach is impractical due to the significant additional computational costs involved. An alternative way is to apply transfer learning techniques to a previously trained model, but this approach is hindered by catastrophic forgetting~\citep{kirkpatrick2017overcoming, mccloskey1989catastrophic}, which leads to degraded performance on earlier classes. In centralized settings, such challenges have been extensively studied within the framework of continual learning (CL)~\citep{zenke2017continual, parisi2019continual, yoon2018lifelong}, where various algorithms have been developed to mitigate catastrophic forgetting from multiple perspectives.



Despite these advancements, most continual learning methods cannot be directly adapted to the federated learning environment due to the essential disparities between the two frameworks. For instance, experience replay~\citep{shin2017continual}—a widely-used strategy that involves storing a subset of past data to preserve some knowledge of previous distributions during training—poses privacy concerns and is not suitable for FL. Therefore, recent studies~\citep{babakniya2024data, qi2023better, ma2022continual} have introduced Federated Continual Learning (FCL), a paradigm addressing catastrophic forgetting in FL environments experiencing evolving data classes. A common scenario in FCL involves the dynamic integration of data with new classes into local clients, a process known as Federated Class Incremental Learning (FCIL)~\citep{dong2022federated, ijcai2023p443}. FCIL enables local clients to continuously gather new data with new classes at any time.




Existing FCIL methods primarily depend on either an unlabeled surrogate dataset to facilitate FL training~\citep{ma2022continual} or require a memory buffer for storing historical data~\citep{dong2022federated}, which are not suitable in privacy-sensitive FL environments, such as hospitals or banks where long-term data retention is discouraged or prohibited. An alternative approach involves utilizing data-free techniques that do not require real-data storage and employing the generative adversarial network (GAN)~\citep{goodfellow2014generative} to simulate historical data. These approaches~\citep{babakniya2024data, zhang2023target} have attracted considerable attention and have been proven effective in mitigating the issue of catastrophic forgetting in FCIL. However, the effectiveness of these strategies is frequently compromised by the inherent instability and high sensitivity of GANs~\citep{wiatrak2019stabilizing}. Such instability impairs their capability to develop a robust global model in FCIL. Compared to GANs, diffusion models~\citep{croitoru2023diffusion} offer distinct advantages including precise control over the generation process, an interpretable latent space, robustness against overfitting, and enhanced stability~\citep{yu2024cross}. Consequently, diffusion models are capable of producing images of superior quality. 




In this paper, we propose a \textbf{D}ata-\textbf{F}ree F\textbf{ed}erated Class Incremental Learning framework with
\textbf{D}iffusion-based \textbf{G}enerative \textbf{M}emory (\method) that employs the diffusion model to generate stable, high-quality images to mitigate the catastrophic forgetting issue in FCIL. We design a new balanced sampler to help train the diffusion models to alleviate the common non-IID~\citep{li2022federated} problem in FL. Additionally, since employing the diffusion model would not always produce higher-quality images with accurate labels which results in performance degradation, we introduce a novel entropy-based sample filtering approach from an information theory perspective to filter out low-confidence samples. Specifically, entropy in information theory quantifies the average level of uncertainty, making it an appropriate criterion for filtering uncertain samples. To this end, we achieve this goal by removing the generative replay samples with low entropy values while preserving those with high entropy values.


Finally, we propose integrating traditional knowledge distillation~\citep{hinton2015distilling} with a feature-based regularization term to enhance knowledge transfer from previous tasks to the new task while minimizing the feature drift from earlier tasks. The proposed feature distance loss addresses a common issue in existing continual learning methods, which typically focus on minimizing the KL divergence in the prediction space between the teacher and student model~\citep{VCL2} while ignoring the features that contain rich semantic information. To the best of our knowledge, this is the first work to explore the diffusion model in FCIL, providing a novel way to address catastrophic forgetting in this field. With the help of the diffusion model, our approach does not require historical samples of previous tasks or external datasets. This data-free approach is particularly useful in scenarios where data sensitivity is a concern. Our framework does not require any additional communication costs compared with the baseline method FedAvg~\citep{mcmahan2017communication}. The efficacy and efficiency of our framework are substantiated through the extensive experimental results of our paper.


We summarize our contributions as follows:

\begin{itemize}
    \item We propose a novel data-free federated class incremental learning framework \method\ to alleviate the catastrophic forgetting issue in FCIL by incorporating diffusion-based generative memory. Our framework incurs no additional communication costs compared to the baseline method FedAvg.
    \item We design a novel balanced sampler to assist in the training of diffusion models to address the prevalent non-IID problem in federated learning.
    \item We introduce a novel entropy-based sample filtering approach to remove lower-quality generative samples from an information theory perspective.
    \item We conduct extensive experiments on three datasets with various non-IID and task settings to show the efficacy of our method.  Our approach consistently outperforms existing FCIL methods, e.g., over a 4\% improvement in average accuracy on the Tiny-ImageNet dataset.
\end{itemize}


\section{Related Work}

\textbf{Continual Learning.} Continual learning (CL)~\citep{zenke2017continual, lesort2020continual, pan2020continual, lopez2017gradient} has been extensively studied in recent years to address the problem of catastrophic forgetting~\citep{mccloskey1989catastrophic, kirkpatrick2017overcoming, lee2017overcoming} in machine learning.  In the CL framework, the training data is presented to the model as a sequence of datasets, commonly referred to as \textbf{tasks}. At each time step, the model has access to only one dataset (task) and seeks to perform well on both current and previous tasks. There are primarily three incremental learning (IL) settings in CL: Task-IL, Domain-IL, and Class-IL. In Task-IL~\citep{masana2021ternary, oren2021defense}, tasks are distinct with separate output spaces identified by task IDs during training and inference phases. In contrast, Domain-IL~\citep{mirza2022efficient, kiyasseh2021clinical} maintains a consistent output space across tasks without the provision of task IDs. Class-IL \citep{mittal2021essentials, masana2022class, belouadah2019il2m} represents a more complex scenario in which each new task introduces additional classes to the output space, progressively increasing the total number of classes. In this study, we focus on class incremental learning (Class-IL), due to its greater relevance to real-world applications.

\textbf{Federated Learning.} Federated learning (FL) is a distributed learning framework that constructs a global model on a central server by aggregating parameters that are independently learned from private data on multiple client devices. FedAvg~\citep{mcmahan2017communication} is a popular FL method whose performance is limited due to the dispersed nature of the data (non-IID). Many FL methods aim to mitigate the issue of data heterogeneity by refining the training process to enhance the global model, such as FedProx~\citep{li2020federatedprox}, FedDisco~\citep{ye2023feddisco}, FedDC~\citep{gao2022feddc}, FedNP~\citep{wu2023fednp}, and CCVR~\citep{luo2021no}. Several methods involve training generative models using distributed resources to enhance model performance \citep{zhang2022dense, zhang2021training}. In this study, we tackle a more complex situation involving statistical heterogeneity in federated learning where users' local data changes over time.

\textbf{Federated Continual Learning.} A few studies have explored continual learning within a federated learning framework. One pioneering work, FedWeIT~\citep{yoon2021federated}, focuses on the Task-IL setting, which requires task IDs during FL training and inference. For federated class incremental learning (FCIL), FLwF2T~\citep{usmanova2021distillation} and GLFC~\citep{dong2022federated} have introduced distillation-based methods to address catastrophic forgetting from both local and global perspectives, respectively. Similarly, CFeD~\citep{ma2022continual} applies knowledge distillation on both the server and client sides using a surrogate dataset to alleviate forgetting. More recently, FedET~\citep{liu2023fedet} utilizes an enhanced transformer to facilitate the absorption and transfer of new knowledge for NLP tasks. Moreover, TARGET~\citep{zhang2023target}, MFCL~\citep{babakniya2024data}, and FedCIL~\citep{qi2023better} all employ generative models to create synthetic data from previous tasks to reduce forgetting, with their efficacy independently confirmed in their respective papers. In this paper, we aim to improve the global model by generating more stable and high-quality synthetic images using diffusion models.

\section{Problem Definition}

\textbf{Federated Learning (FL).} Assume there are $N$ different clients, each client $i$ has its own private dataset $D_i = \{(\mathbf{x}_k, y_k)\}_{k=1}^{n_i}$ with size $n_i$. The goal of federated learning is for clients to collaboratively develop a global model $\theta_g$ without sharing their private datasets:

\begin{equation}
   \theta_g =  \mathop{\argmin}\limits_{\theta} \sum_{i=1}^{N} \mathbb{E}_{(\mathbf{x},y)\sim {\mathbb  P}_{D_i}}[L(\theta; \mathbf{x}, y)],
\end{equation}

\noindent where $L(\theta; \mathbf{x}, y)$ is the loss function evaluated on a dataset $D= \{(\mathbf{x}, y)\}$ with model $\theta$, and ${\mathbb  P}_{D_i}$ is the data distribution of private dataset ${D_i}$. 

\textbf{Class Incremental Learning (Class-IL).} Consider a set of $T$ tasks $\mathcal{T} = \{\mathcal{T}^t\}_{t=1}^{T}$, each task $\mathcal{T}^t$ contains a dataset $D^t=\{(\mathbf{x}_k^t, y_k^t)\}_{k=1}^{n^t}$ with size $n^t$. Every new task $\mathcal{T}^t$ introduces $n^t_\mathcal{C}$ new classes $\mathcal{C}^t=\{c_j^t\}_{j=1}^{n^t_\mathcal{C}}$ that are not present in the previous tasks. During the training of task $\mathcal{T}^t$, datasets from all previous tasks $\{D^i | i < t\}$ are inaccessible. The objective of class incremental learning is to develop a model $\theta^T$ that not only performs effectively on the last task $\mathcal{T}^T$, but also preserves its performance on all previous tasks $\{\mathcal{T}^t\}_{t=1}^{T-1}$ to avoid catastrophic forgetting:

\begin{equation}
\theta^{T} = \argmin_{\theta} \sum_{t=1}^{T} \mathbb{E}_{(\mathbf{x},y)\sim {\mathbb P}_{D^t}}[L(\theta; \mathbf{x}, y)], 
\end{equation}


\textbf{Federated Class Incremental Learning (FCIL).} FCIL incorporates both federated learning and class incremental learning principles to develop a global model that can incrementally learn new classes across multiple distributed clients while maintaining data privacy. Within this framework, we have in total $T$ tasks $\mathcal{T} = \{\mathcal{T}^t\}_{t=1}^{T}$, each client $i$ will independently learn from $T$ local tasks $\mathcal{T}_i = \{\mathcal{T}^t_i\}_{t=1}^{T}$ in a class-incremental way, and $\mathcal{T}^t = \cup_{i=1}^{N}\mathcal{T}^t_i$. At a given step $t$, client $i$ will only have access to dataset $D_i^t$ from task $\mathcal{T}^t_i$. During training, clients will communicate with the central server to update and obtain the global model. The objective of FCIL is to develop a comprehensive global model, $\theta_g^T$, that effectively predicts all classes from all $T$ tasks encountered by all $N$ clients:


\begin{equation}
\theta_g^T = \argmin_{\theta} \sum_{i=1}^{N} \sum_{t=1}^{T} \mathbb{E}_{(\mathbf{x},y)\sim {\mathbb P}_{D_i^t}}[L(\theta; \mathbf{x}, y)], 
\end{equation}


\section{Federated Class Incremental Learning with \method}


\label{sec:method}

\begin{figure}
    \centering
    \includegraphics[width=1\linewidth]{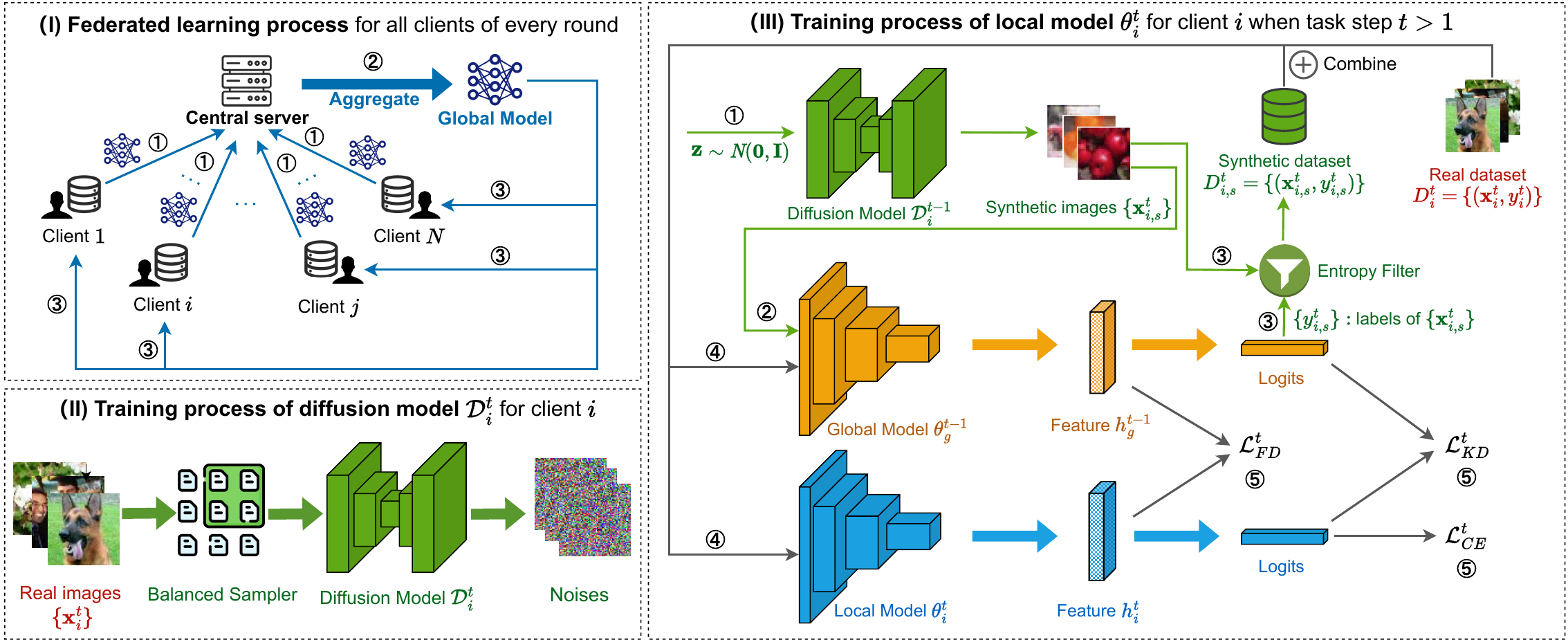}
    \caption{Overview of our framework. A diffusion model is trained with the help of the \textit{Balanced Sampler} (II). We utilize the diffusion model to generate synthetic images, and assign labels to these images by the global model $\theta_g^{t-1}$ from the previous task. An \textit{Entropy Filter} subsequently screens these samples. We combine three losses ($\mathcal{L}_{CE}^t, \mathcal{L}_{KD}^t$ and $\mathcal{L}_{FD}^t$) to train the local model $\theta_i^t$ (III). }
    \label{fig:overview}
\end{figure}

In FCIL, storing previous data is considered a violation of the FL setup and is thus prohibited. In this paper, we propose a data-free framework \textit{\method}\ which employs diffusion models to generate synthetic datasets to tackle the catastrophic forgetting problem in FCIL while adhering to the data privacy standards of FL. 
The overview of our framework is shown in Figure \ref{fig:overview}.

Previous studies focused on training a GAN on the server side to generate synthetic data. However, the quality of the generated images tends to be unstable and highly sensitive to the training hyperparameters due to the inherent limitations of GAN~\citep{alqahtani2021applications}. Moreover, the model-inversion technique necessitates pre-assigned labels for image generation, rather than utilization of labels predicted by the global model. This method often fails to accurately capture the true distribution or nuances of the original data~\citep{he2020towards}. Consequently, it is advisable to train a generator directly on the client side to produce higher-quality images. The global model will then label these images, thereby creating synthetic samples for subsequent tasks.


Diffusion models offer several advantages~\citep{yu2024cross} over GANs, including improved control during the generation phase, increased robustness to overfitting, and enhanced stability. In this study, we propose training a diffusion model subsequent to local training on every client $i$ using dataset $D_i^t$ from the current task $\mathcal{T}^t_i$ to assist future task training. We first introduce how to train diffusion models in Section \ref{sec:dm}, then elaborate on the local training process using the diffusion model in Section \ref{sec:lt}.



\subsection{Train Diffusion Models with \textit{Balanced Sampler}}
\label{sec:dm}




Due to the pervasive non-IID issue in federated learning (FL) \citep{li2020federatedprox}, employing the traditional random sampling method from dataset $D_i^t$ to train the diffusion model is inadvisable. This method could lead to the generation of images with a highly unbalanced distribution. In extreme cases, it is imaginable that images from certain classes might not be generated at all. This imbalance may lead the training of the local model to diverge from the global optimum. Therefore, it is essential to develop a training strategy for the diffusion model to address the non-IID issues during image generation.


\begin{algorithm}[t]
    \begin{algorithmic}[1]
    \renewcommand{\algorithmicrequire}{\textbf{Input:}}
	\renewcommand{\algorithmicensure}{\textbf{Output:}}
	\caption{Training procedure for the diffusion model with \textit{Balanced Sampler}}
	\label{alg1}
        \REQUIRE Local dataset $D_i^t$ of task $\mathcal{T}^t_i$ on client $i$, diffusion model $\mathcal{D}_i^{t-1}$ from task $\mathcal{T}^{t-1}_i$ of client $i$, local training epochs $E_\mathcal{D}$, batch size $B$, number of new classes $n^t_i$ for task $\mathcal{T}^t_i$ on client $i$.
        \ENSURE  Diffusion model $\mathcal{D}_i^t$.
        
        \STATE $ \mathcal{D}_i^{t} \leftarrow  \mathcal{D}_i^{t-1}$ // For the first task $t=1$, random initialize $\mathcal{D}_i^{1}$ 
        \STATE $B_\mathcal{C} \leftarrow \lceil \frac{B}{n^t_i} \rceil $ // Number of samples selected from every class for every batch
        \STATE $N_{\mathcal{C}} \leftarrow \max \{|D_i^t\textbf{[}c_j^t\textbf{]}|\}_{j=1}^{n^t_i}$ // The sample numbers of the class with the most samples
        \STATE $E_B \leftarrow \lceil \frac{N_{\mathcal{C}}}{B_{\mathcal{C}}} \rceil$ // Total number of batches for every epoch
        \FOR{local epoch $e= 1, 2, \ldots, E_{\mathcal{D}}$}
               \FOR{class $c_j^t=1, 2, \ldots, n^t_i$}
                    \STATE $\mathcal{S}_j^t \leftarrow \mathcal{X} (D_i^t\textbf{[}c_j^t\textbf{]})$ // $\mathcal{X}(D)$ means select the images part of $D$
                    \FOR{$l=1, 2, \cdots, \lceil \frac{N_{\mathcal{C}}}{|\mathcal{S}_j^t|} \rceil-1$}
                        \STATE $\mathcal{S}_j^t \leftarrow \mathcal{S}_j^t \cup \textit{Reshuffle}(\mathcal{X} (D_i^t\textbf{[}c_j^t\textbf{]}))$
                    \ENDFOR
                \ENDFOR
            \FOR{batch $b= 1, 2, \ldots, E_B$}
                \STATE $\mathcal{S} \leftarrow \emptyset$ // Sample set for current batch
                \FOR{class $c_j^t=1, 2, \ldots, n^t_i$}
                        \STATE $\mathcal{S} \leftarrow \mathcal{S} \cup \mathcal{S}_j^t\textbf{[}(b-1)\times B_\mathcal{C}:b\times B_\mathcal{C}\textbf{]}$
                \ENDFOR
                \STATE $\mathcal{D}_i^{t} \leftarrow$ \textit{Update}$(\mathcal{D}_i^{t}, \mathcal{S})$ // Train diffusion model with sample set $\mathcal{S}$ by method like \textit{DDPM}
            \ENDFOR
        \ENDFOR
    \end{algorithmic}
\end{algorithm}

As shown in Algorithm \ref{alg1}, we have designed a mechanism named \textit{Balanced Sampler} to help train the diffusion model, enabling it to generate more balanced images across different classes. Due to the non-IID nature of FL, the number of samples for each class on a client can vary significantly. Therefore, instead of randomly selecting $B$ samples from all samples for each batch, our approach involves selecting an equal number of $B_\mathcal{C}$ samples from each class to form a training data batch. This strategy allows the diffusion model to consistently learn from an equitable distribution of samples across all classes in every iteration, effectively addressing the challenges posed by non-IID data.



In detail, we generate $E_B$ batches of samples per training epoch. Every batch consists of $B_\mathcal{C}$ randomly selected samples from each class, leading to a total of $B$ samples per batch (Lines 2-4). However, since the number of samples is imbalanced, there will be a batch $b$ in which all samples of a specific class $c_j^t$ are selected in earlier batches, leaving some samples from other classes unselected. In such cases, we reshuffle the samples from class $c_j^t$ and start a cycle of new random selection for class $c_j^t$ into the batch. This process of reshuffling is repeated until all samples from the class with the largest sample size have been completely incorporated for one training epoch (Lines 6-18). This sampling strategy ensures that the diffusion model consistently interacts with samples from every class in each training batch, thereby producing more class-balanced images.


All the available training methods for diffusion models, such as DDPM~\citep{ho2020denoising} and DDIM~\citep{song2021denoising}, are applicable for training our generative model. Furthermore, the diffusion model $\mathcal{D}_i^t$ is initialized with the model $\mathcal{D}_i^{t-1}$, which was trained on the previous task $\mathcal{T}^{t-1}_i$ (Line 1), enabling it to retain the data distributions from previous tasks. This allows the model to remember and generate samples for both the current and previous tasks, thus effectively addressing the issue of catastrophic forgetting.


\subsection{Entropy Filter for Selecting High-Quality Generative Replay Samples}
\label{sec:lt}

In our framework, client $i$ will utilize its local dataset $D_i^1$ to train the local model $\theta^1_i$ at the first learning task  $\mathcal{T}^1_i$. For subsequent tasks $\{\mathcal{T}^t_i | t>1\}$, the diffusion model $\mathcal{D}_i^{t-1}$ trained from the previous task $\mathcal{T}^{t-1}_i$ will generate synthetic data to help the client in relieving catastrophic forgetting. 


Given random noises $\textbf{z} \sim \textbf{\textit{N}}({\bf 0},{\bf I})$, the number of samples to be generated $n_s$, and the retention ratio $\lambda$ of the \textit{Entropy Filter} (to be discussed later), we can utilize existing sampling methods, such as DDPM~\citep{ho2020denoising} or DDIM~\citep{song2021denoising}, to generate $\frac{n_s}{\lambda}$ synthetic images $\{\mathbf{x}_{i,s}^t\}$ with the diffusion model $\mathcal{D}_i^{t-1}: \{\mathbf{x}_{i,s}^t\} \leftarrow \mathcal{D}_i^{t-1}(\textbf{z}, \frac{n_s}{\lambda})$. Since diffusion models produce images without labels, it is necessary to assign labels to these generated images for subsequent training.



 Given that the global model generally outperforms local models on clients in FL, we utilize the global model $\theta_g^{t-1}$ to annotate the generated images on the clients: $\{y_{i,s}^t\} = \{\theta_g^{t-1}(\mathbf{x}_{i,s}^t)\}$. However, the global model does not always exhibit high confidence in the labels assigned to certain images. Since labels are determined by selecting the class with the highest probability after applying the softmax function to the logits from the final layer of the model, there are cases where a class is chosen as the label simply because its probability is slightly higher than that of other classes. These low-confidence labels often do not accurately match the generated images, leading to fake samples that impair training performance. Hence, it is essential to design a method to filter out such data.



    To address the above issue, we propose an \textit{Entropy Filter}, inspired by information theory. According to information theory, entropy measures the amount of information conveyed by an event, whereby higher entropy corresponds to greater information value. For each image, we calculate the entropy of the probability distribution across all classes to evaluate the global model's confidence in the assigned label. A higher entropy value indicates greater confidence in the image and we should maintain these samples. The entropy of the predictions achieved by the global model is defined as:
  \begin{equation}
     H(\theta_g^{t-1}(\mathbf{x})) = -\sum\nolimits_{k=1}^{t-1}\sum\nolimits_{j=1}^{n_\mathcal{C}^k} p(c_j^k) \log p(c_j^k), 
  \end{equation}

  where $p(c_j^k)$ is the probability of class $c_j^k$ calculated by global model $\theta_g^{t-1}$. We calculate the entropy of all generated images and sort them in descending order. We then retain only the top $\lambda$ portion of the samples, thereby filtering out samples for which the global model exhibits low confidence:
\begin{equation}
    D^t_{i, s}=\{(\mathbf{x}_{i,s}^t, y_{i,s}^t) \mid i \in \mathcal{I}_\lambda \}, \mathcal{I}_\lambda = \{ i \mid i \leq \lambda \cdot 
    |\{H(\theta_g^{t-1}(\mathbf{x}_{i,s}^t))\}| \},
\end{equation}

where $|\{H(\theta_g^{t-1}(\mathbf{x}_{i,s}^t))\}|$ is the original number of generated images ($\frac{n_s}{\lambda}$). The refined generated dataset $D^t_{i, s}$ is then combined with the real dataset $D^t_{i}=\{(\mathbf{x}_{i}^t, y_{i}^t)\}$ for local model training.
    

  

 

\subsection{The Final Objective Function}

Following the standard FL paradigm, we first apply the \textit{Cross-Entropy (CE)} loss function to minimize the divergence between the predicted and true distributions of new classes in task $\mathcal{T}^t_i$ from dataset $D^t_{i}$, as well as the true distribution of old classes across task $\mathcal{T}^1_i$ to $\mathcal{T}^{t-1}_i$ with dataset $D^t_{i, s}$:

\begin{equation}
    \mathcal{L}_{CE} = CE(\theta^t_i(\mathbf{x}),y;D_i^t, D^t_{i, s}).
\end{equation}

However, solely applying cross-entropy is insufficient to effectively address the problem of forgetting. To address this issue, we propose to use \textit{Knowledge Distillation}~\citep{hinton2015distilling} to transfer knowledge from the previous global model $\theta^{t-1}_g$ to the current local model $\theta^t_i$, thereby aligning the output distributions of the two models. It is essential when dealing with continuously changing data distributions, as it enhances the model's robustness and generalization capabilities across various tasks:

\begin{equation}
    \mathcal{L}_{KD} = KL(\theta^t_i(\mathbf{x}),\theta^{t-1}_g(\mathbf{x});D_i^t, D^t_{i, s}),
\end{equation}

where $KL$ is the Kullback-Leibler divergence between the outputs of the student model $\theta^t_i(\mathbf{x})$ and the teacher model $\theta^{t-1}_g(\mathbf{x})$ over the data distributions from $D_i^t \cup D^t_{i, s}$. 


Finally, to ensure that the most significant features with richer semantic information from old models are transferred, while allowing the model to learn new features from new tasks better~\citep{smith2021always}, we introduce an additional \textit{Feature Distance} loss to control the drift of feature distribution:


\begin{equation}
    \mathcal{L}_{FD} = ||h_i^t(\mathbf{x};D_i^t, D^t_{i, s})- h_g^{t-1}(\mathbf{x};D_i^t, D^t_{i, s})||^2_2,
\end{equation}

where $|| \cdot  ||^2_2$ denotes the L2 distance. $h_i^t(\mathbf{x}; D_i^t, D^t_{i, s})$ and $h_g^{t-1}(\mathbf{x}; D_i^t, D^t_{i, s})$ are feature representations from the feature extractors of the current local model and the previous global model. Minimizing this loss ensures the features remain stable over time in the model's feature space, which is crucial for tasks requiring consistent feature interpretations amidst continuously evolving data distributions.

To summarize, the total loss function for model $\theta_i^t$ is formed as follows:

\begin{equation}
    \mathcal{L}(\theta_i^t) = \mathcal{L}_{CE} + \alpha \mathcal{L}_{KD} + \gamma \mathcal{L}_{FD},
\end{equation}

where $\alpha$ and $\gamma$ are two hyperparameters controlling the effect of both losses on model training. The values of $\alpha$ and $\gamma$ are determined empirically in the experiments. 


\subsection{Model Aggregation and Analysis}

In our framework, diffusion models on local clients remain private and are not shared to avoid potential privacy concerns and reduce communication costs. The server simply collects and aggregates the local models from clients to produce a comprehensive global model. Our approach greatly diminishes communication overhead compared to studies that create a synthetic dataset on the server and distribute it to clients. The pseudocode of our framework is provided in Algorithm \ref{alg2} in the appendix.

\textbf{Communication Cost.} In our method, each client sends only their local model to the central server. Thus, the overall communication cost for $T$ tasks on $N$ clients is $O(TNRM)$, where $R$ is the communication rounds per task and $M$ is the size of the local model, consistent with FedAvg~\citep{mcmahan2017communication}.

\section{Experiments}
\subsection{Experimental Setup}

\textbf{Datasets and Settings.} We evaluate the \textit{Average Accuracy} ($Acc$) and \textit{Average Forgetting} of different methods with three challenging datasets: \textbf{EMNIST-Letters}~\citep{cohen2017emnist}, \textbf{CIFAR-100}~\citep{krizhevsky2009learning}, and \textbf{Tiny-ImageNet}~\citep{le2015tiny}, consisting of 26, 100, and 200 classes, respectively. By default, each dataset is partitioned into $T=5$ non-overlapping tasks and distributed among $N=10$ clients using a Dirichlet distribution~\citep{ng2011dirichlet} with $\beta=0.5$. The clients train local models with the ResNet-18~\citep{he2016deep} structure. For the diffusion model, we employ the same architecture as DDPM~\citep{ho2020denoising} on the clients. 


\textbf{Baselines.}
We compare our method with the following baselines: \textbf{FedAvg}~\citep{mcmahan2017communication}, \textbf{FedProx}~\citep{li2020federatedprox}, 
\textbf{FedEWC}~\citep{lee2017overcoming},
\textbf{FLwF-2T}~\citep{usmanova2021distillation}, \textbf{FedCIL}~\citep{qi2023better}, \textbf{TARGET}~\citep{zhang2023target}, and \textbf{MFCL}~\citep{babakniya2024data}. \textbf{FedAvg} and \textbf{FedProx} are two representative methods in federated learning, which we have adapted to FCIL setting; \textbf{FedEWC} is the federated adaptation of the incremental moment matching approach commonly used in continual learning; \textbf{FLwF-2T} utilized knowledge distillation to address catastrophic forgetting in FCIL; \textbf{FedCIL} is a generative replay method for clients to train the ACGAN~\citep{odena2017conditional} to produce synthetic samples from previous tasks. Lastly, \textbf{TARGET} and \textbf{MFCL} are two recent generator-based methods to invert images from the global model on the server with different types of losses.

For more implementation details about the experimental setup, such as the details of metrics, optimizer, learning rate and batch size we employed, please refer to Section \ref{sec:setup_supp} of the appendix.

\subsection{Performance of \method}

\begin{table}[t]
    \centering
    \caption{Performance (\%, mean$\pm$std) comparison of our method to other baselines on $T$=5 tasks.}
    \scalebox{0.87}{
\begin{tabular}{c|cc|cc|cc}
\myhline
Dataset & \multicolumn{2}{c|}{EMNIST-Letters} & \multicolumn{2}{c|}{CIFAR-100} & \multicolumn{2}{c}{Tiny-ImageNet}  \\
\hline
Metric & Acc ($\uparrow$)   &  $\mathcal{F}$ ($\downarrow$)   & Acc ($\uparrow$)   &  $\mathcal{F}$ ($\downarrow$)  & Acc ($\uparrow$)   &  $\mathcal{F}$ ($\downarrow$) \\
\hline
\textbf{FedAvg} & 45.41$\pm$0.49 & 77.48$\pm$0.68 & 38.42$\pm$0.44 & 61.65$\pm$0.81 & 24.53$\pm$1.44 & 42.75$\pm$0.98 \\
\textbf{FedProx} & 46.83$\pm$0.92 & 79.42$\pm$1.33 & 38.48$\pm$0.49 & 61.03$\pm$0.44 & 24.64$\pm$1.02 & 43.18$\pm$1.70 \\
\hline
\textbf{FedEWC} & 62.92$\pm$0.87 & 70.46$\pm$0.66 & 40.74$\pm$0.54 & 56.83$\pm$0.68 & 25.38$\pm$1.28 & 36.27$\pm$1.16 \\
\textbf{FLwF-2T} & 76.44$\pm$0.95 & 32.26$\pm$0.84 & 50.43$\pm$0.18 & 33.28$\pm$0.66 & 26.02$\pm$0.62 & 31.64$\pm$0.53 \\
\hline
\textbf{FedCIL} & 79.59$\pm$0.43 & 31.07$\pm$0.29 & 53.18$\pm$0.74 & 32.66$\pm$0.29 & 27.45$\pm$1.74 & 30.02$\pm$0.58 \\
\textbf{TARGET} & 77.36$\pm$0.59 & 28.83$\pm$0.37 & 54.01$\pm$0.56 & 29.41$\pm$0.73 & 27.88$\pm$0.89 & 29.83$\pm$0.76 \\
\textbf{MFCL}  & 80.21$\pm$0.44 & 29.25$\pm$0.58 & 48.87$\pm$0.26 & 31.47$\pm$0.67 & 29.08$\pm$0.73 & 27.61$\pm$0.63 \\
\textbf{\method\ (Ours)}  & \textbf{85.33$\pm$0.36} & \textbf{26.34$\pm$0.28} & \textbf{57.08$\pm$0.58} & \textbf{27.64$\pm$0.74} & \textbf{33.55$\pm$0.69} & \textbf{23.54$\pm$0.72} \\
\myhline
\end{tabular}%
}

    \label{tab:main_results}
\end{table}%

\textbf{Main results.} Table~\ref{tab:main_results} presents the performance of different methods. Results show that our \method\ method consistently outperforms all other methods across all datasets. Specifically, \method\ exceeds existing methods by over 5\% in average accuracy ($Acc$) on the EMNIST-Letters dataset, and reduces average forgetting ($\mathcal{F}$) by more than 4\% on the Tiny-ImageNet dataset. Notably, FedAvg and FedProx exhibit the highest levels of forgetting as they are not designed for FCIL. FedEWC, which employs regularization constraints during training, failed to prevent forgetting due to insufficient data availability. In contrast, distillation-based approaches, including FLwF-2T, TARGET, MFCL, and our method, effectively improve average accuracy and mitigate catastrophic forgetting by transferring knowledge from old to new tasks. This underscores the efficacy of knowledge distillation.

On the other hand, generative methods like FedCIL, TARGET, MFCL, and our proposed method significantly enhance accuracy and mitigate forgetting by generating synthetic datasets for future tasks. This underscores the capacity of synthetic data to accurately reflect the distribution of historical data. Among these approaches, our method exhibits the least forgetting and the highest accuracy, substantiating its effectiveness in preserving knowledge from previous tasks.

\begin{figure}[t]
    \centering
    \includegraphics[width=0.9\linewidth]{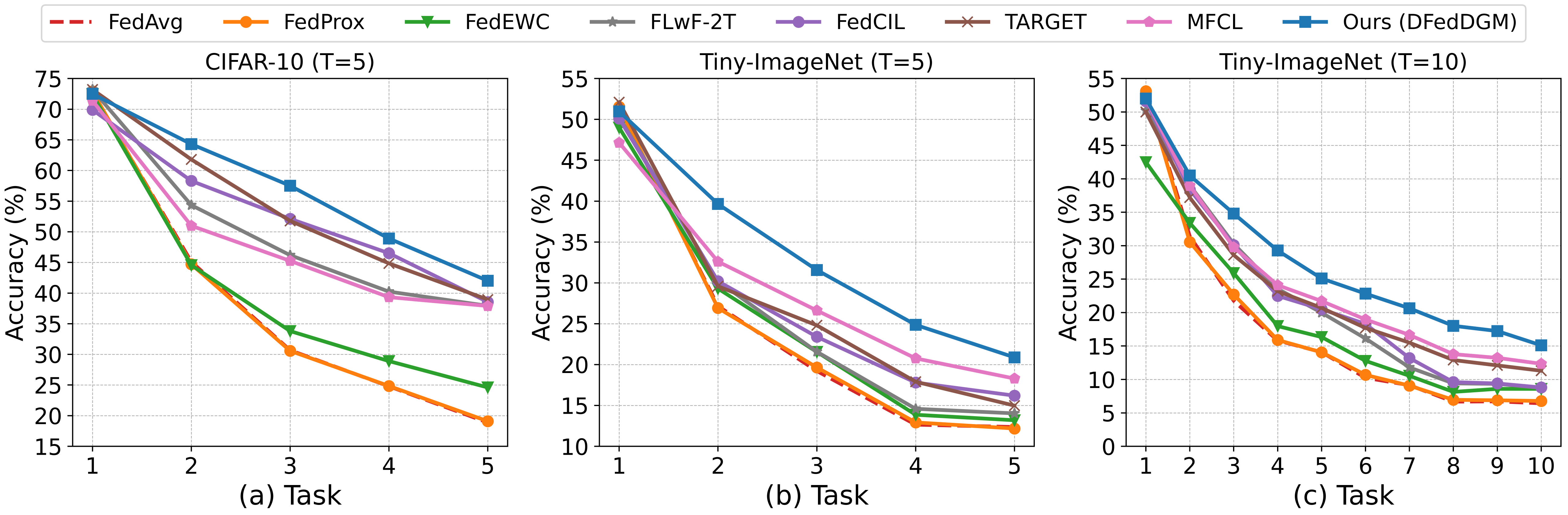}
    \caption{Test average accuracy vs. the number of observed tasks for (a) CIFAR-100 on $T=5$ tasks, (b) Tiny-ImageNet on $T=5$ tasks, (c) Tiny-ImageNet on $T=10$ tasks.}
    \label{fig:tasks}
\end{figure}

\textbf{Effect of the number of tasks.} Figure \ref{fig:tasks} depicts the model's performance on both current and prior tasks after the completion of each task. The curves indicate that our proposed model consistently surpasses other baseline methods in all incremental tasks, irrespective of the number of tasks (whether $T=5$ or $T=10$). This result highlights the efficacy of our approach in enabling multiple local clients to learn new classes sequentially while alleviating the issue of forgetting.

\begin{figure}[t]
  \centering
  \begin{subfigure}{0.32\linewidth}
    \includegraphics[width=\linewidth]{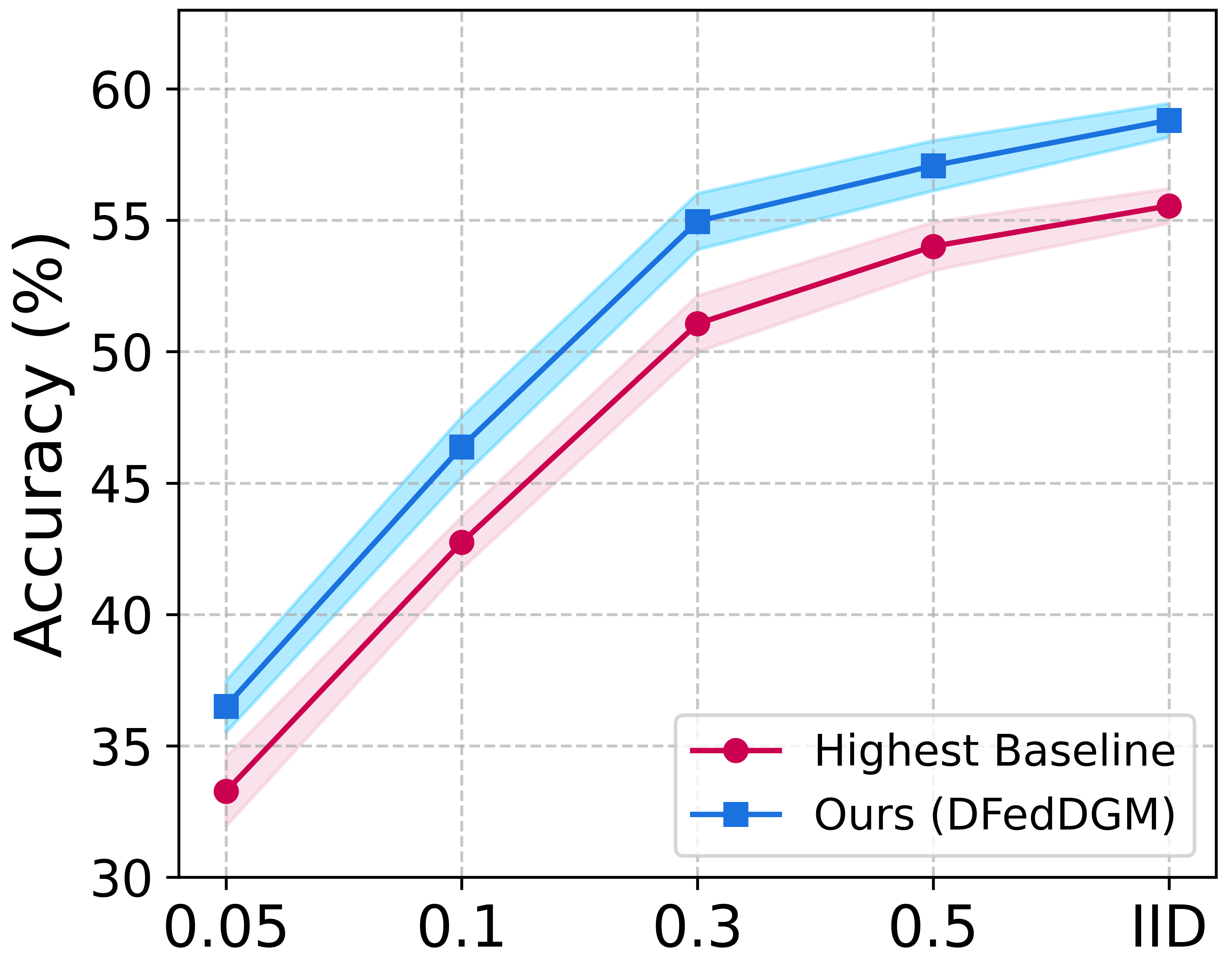}
    \caption{Non-IIDness}
  \end{subfigure}
  \hspace{0.5mm}
  \begin{subfigure}{0.64\linewidth}
    \includegraphics[width=\linewidth]{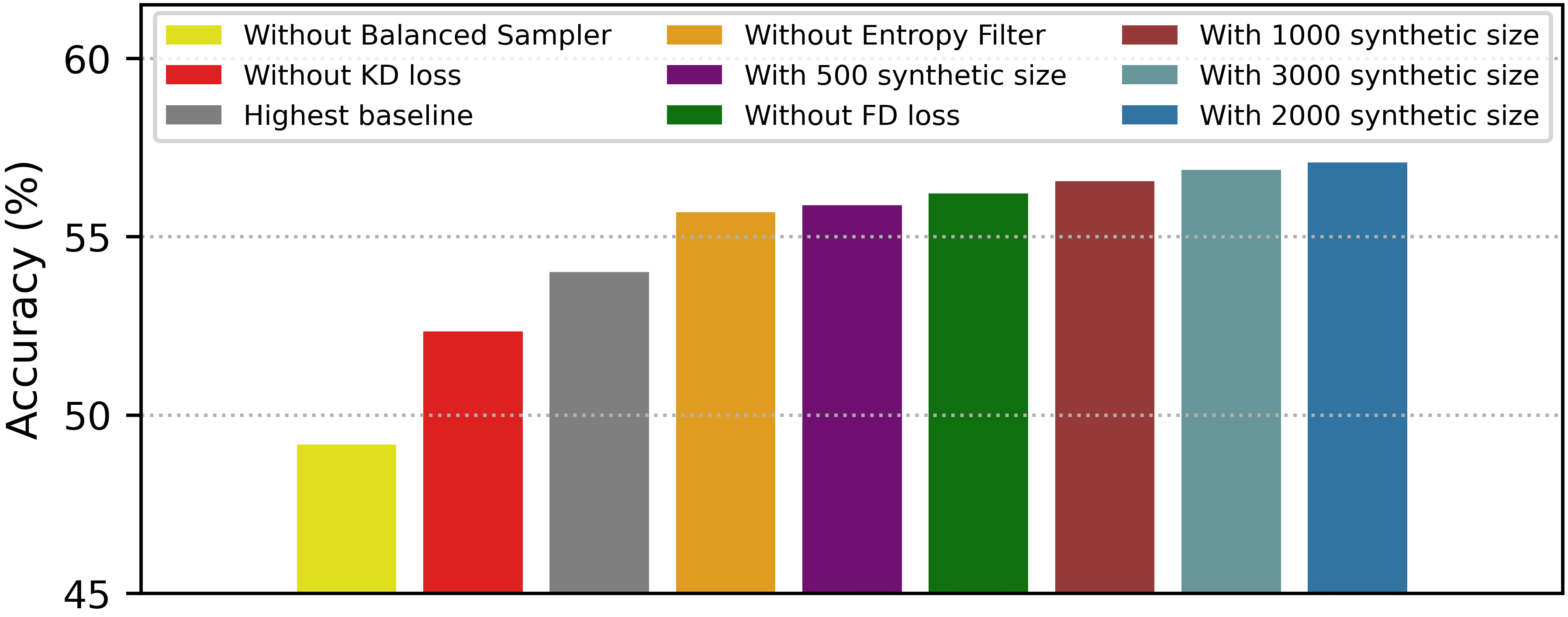}
    \caption{Ablation method}
  \end{subfigure}
\caption{(a) Average accuracy under different data distribution settings on the CIFAR-100 dataset, (b) Ablation studies on the CIFAR-100 dataset.} 
   \label{fig:ablation}
\end{figure}

\textbf{Impact of data distribution.} We performed experiments across various non-IID settings by adjusting the Dirichlet parameters to $\{$0.05, 0.1, 0.3, 0.5$\}$, as well as the IID setting. Figure \ref{fig:ablation} (a) shows that our approach consistently outperforms the \textit{Highest Baseline}, which is the highest accuracy achieved by any of the baseline methods, across all non-IID settings. This highlights the notable efficacy of our approach in alleviating catastrophic forgetting, even when faced with extreme data distributions.





\subsection{Ablation Studies}


In Figure \ref{fig:ablation} (b), we illustrate the significance of each component in our framework. We can see that the \textit{Balanced Sampler} is crucial for the performance, underscoring the necessity of generating balanced samples to mitigate the effects of non-IID data. Removing the \textit{Entropy Filter} leads to a performance decline due to the inclusion of excessive synthetic samples with low-confidence labels, highlighting the importance of filtering data based on prediction entropy. Additionally, the knowledge distillation (KD) loss is also vital for performance enhancement, validating its importance in mitigating forgetting; the FD loss also contributes to an accuracy improvement, emphasizing the importance of maintaining temporal consistency within the feature space for continual learning.


Lastly, we evaluate the performance of our method by varying the number of generated samples $n_s$ for the diffusion model. As illustrated in Figure \ref{fig:ablation} (b), the optimal value for $n_s$ is 2000, and performance declines when generating either too few samples (500) or too many samples (3000). Generating too few samples does not provide sufficient synthetic data for the model to learn from previous tasks, while generating too many samples impedes the model's ability to learn new information from the current task. Therefore, striking a balance of $n_s$ is crucial for achieving optimal model performance.
\subsection{Further Analysis}
\label{sec:discuss}

\begin{figure}[t]
    \centering
    \includegraphics[width=0.7\linewidth]{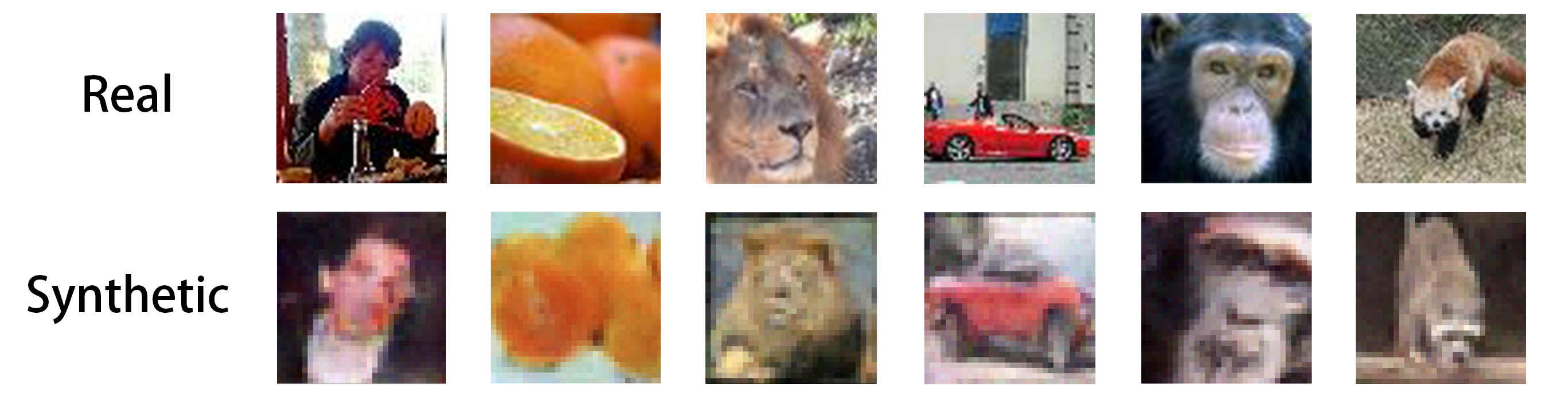} 
    \caption{Real vs synthetic data generated by the diffusion model for the Tiny-ImageNet dataset.}
    \label{fig:syn}
\end{figure}

\textbf{Privacy of \method.} In contrast to existing FCIL methods~\citep{zhang2023target, babakniya2024data}, which require clients to share locally trained generative models or synthetic images, our diffusion model is never shared with others. Consequently, our method does not introduce additional privacy concerns and retains the same privacy protections as FedAvg~\citep{mcmahan2017communication}. Specifically, our method can effectively defend attacks such as model poisoning~\citep{fang2020local}, backdoor attacks~\citep{chen2022defending}, and gradient inversion~\citep{geiping2020inverting} through techniques like secure aggregation~\citep{fereidooni2021safelearn} and differential privacy~\citep{wei2020federated}.


Figure \ref{fig:syn} presents several synthetic images that exhibit notable differences from real data. Nevertheless, they are clear to identify and contain essential class-specific knowledge to accurately represent the entire class, which facilitates knowledge transfer and substantially mitigates catastrophic forgetting.


\textbf{Limitations.} Due to the inherent limitations of the diffusion model~\citep{ho2020denoising}, our method introduces additional computational overhead and extends the training time compared to other approaches. However, unlike other generative methods, our framework does not incur additional communication costs, which is a more critical issue in FL. Compared to other GAN-based methods, our approach produces more stable images and offers significant improvements in model performance. Furthermore, the training time can also be reduced by employing advanced techniques such as the quick sampling method DDIM~\cite{song2021denoising}, which we plan to incorporate in future work to better adapt to the FCIL scenarios. 



\section{Conclusion}


This paper presents a novel framework, \method\, to mitigate the issue of catastrophic forgetting in federated class incremental learning (FCIL). We employ diffusion models to generate high-quality synthetic images on the client side in a data-free manner, and utilize knowledge distillation to efficiently transfer information from previous tasks. Extensive experiments demonstrate the effectiveness of our method compared to existing FCIL frameworks.

\newpage

\bibliographystyle{plainnat}
\bibliography{ref}

\newpage

\appendix

\section{\method\ Algorithm}

\label{sec:appalg}

Algorithm \ref{alg2} summarizes our framework. All clients conduct local training independently and then train their diffusion models locally. The server receives only the local models from clients to perform model aggregation, and redistributes the global model to clients.

\begin{algorithm}
	\renewcommand{\algorithmicrequire}{\textbf{Input:}}
	\renewcommand{\algorithmicensure}{\textbf{Output:}}
	\caption{\method}
	\label{alg2}
	\begin{algorithmic}[1]
        \REQUIRE Local datasets $D = \{D_i\}_{i=1}^{N}$, $D_i=\{D_i^t\}_{t=1}^T$, number of clients selected per round $m$, learning rate $\eta$, local training epochs $E$, FL communication rounds per task $R$, number of new classes $n^t_\mathcal{C}$ in task $\mathcal{T}^t$, number of generated images $n_s$, retention ratio of the entropy filter $\lambda$, training hyperparameters $\alpha$, $\gamma$.
        \ENSURE  Final global model $\theta_g^T$.
        \STATE Initialize global model $\theta_g^1$ of task $\mathcal{T}^1$
        \STATE $n_\mathcal{C} \leftarrow 0$ // Number of observed classes
		\FOR{task step $t= 1, 2, \ldots, T$}
            \STATE $n_\mathcal{C} \leftarrow n_\mathcal{C} + n^t_\mathcal{C}$
            \STATE $\theta_g^t \leftarrow$ \textbf{UpdateStructure}$(\theta_g^t, n_\mathcal{C})$ // Incorporate new classes into the classification layer
            \FOR{round $r= 1, 2, \ldots, R$}
                \STATE $C_r \leftarrow$ \textbf{RandomSelect}($N, m$) // Random select $m$ clients from $N$ clients
                \FOR{client $i \in C_r$ \textbf{in parallel}}
                    \STATE $\theta_i^t \leftarrow$ \textbf{ClientUpdate}$(\theta_g^t, D_i^t, \theta_g^{t-1}, \mathcal{D}_i^{t-1})$ // No need of  $\mathcal{D}_i^0$ and $\theta_g^0$ for $t=1$
                    \STATE $\mathcal{D}_i^t \leftarrow$ \textbf{TrainDiffusionModel}($D_i^t, \mathcal{D}_i^{t-1}$) // Refer to Algorithm \ref{alg1}.
                \ENDFOR
                \STATE $\theta_g^t \leftarrow$ \textbf{GlobalAggregation}$(\{\theta_i^t\}_{i\in{C_r}})$ // Average the local models from selected clients
            \ENDFOR
        \ENDFOR
    
        
        \STATE \textbf{ClientUpdate}$(\theta_g^t, D_i^t, \theta_g^{t-1}, \mathcal{D}_i^{t-1}):$
        \STATE $\theta_i^t \leftarrow \theta_g^t$ // Initialize local model as global model
        \FOR{local epoch $e= 1, 2, \ldots, E$}
        \IF{$t = 1$}
            \STATE $\theta_i^t \leftarrow \theta_i^t - \eta \nabla_\theta \mathcal{L}_{CE}(\theta_i^t;D_i^t)$
        \ELSE
        \STATE $\textbf{z}\sim \textbf{\textit{N}}({\bf 0},{\bf I})$ // Sample noises
        \STATE $\{\mathbf{x}_{i,s}^t\} \leftarrow \mathcal{D}_i^{t-1}(\textbf{z}, \frac{n_s}{\lambda})$ // Generate $\frac{n_s}{\lambda}$ synthetic images by diffusion model
        \STATE $D^t_{i, s}=\{(\mathbf{x}_{i,s}^t, y_{i,s}^t)\} \leftarrow \textbf{EntropyFilter}(\{\theta_g^{t-1}(\mathbf{x}_{i,s}^t)\}, \lambda)$ // Filter out part of generated samples and assign labels by model $\theta_g^{t-1}$ for the rest to form the synthetic dataset $D^t_{i, s}$
        \STATE $\mathcal{L}(\theta_i^t) \leftarrow \mathcal{L}_{CE}(\theta_i^t;D_i^t, D^t_{i, s}) + \alpha \mathcal{L}_{KD}(\theta_i^t, \theta_g^{t-1};D_i^t, D^t_{i, s}) + \gamma \mathcal{L}_{FD}(h_i^t, h_g^{t-1};D_i^t, D^t_{i, s})$
        \STATE $\theta_i^t \leftarrow \theta_i^t - \eta \nabla_\theta \mathcal{L}(\theta_i^t)$
        \ENDIF
        \ENDFOR      
	\end{algorithmic}  
\end{algorithm}

\section{Experimental Setup}
\label{sec:setup_supp}




 \textbf{Evaluation Metrics.} We employ the \textit{Average Accuracy} ($Acc$) and \textit{Average Forgetting} ($\mathcal{F}$) to conduct our evaluations. Let \textit{Accuracy} ($Acc^t$) represent the model's accuracy at the end of task $\mathcal{T}^t$ across all observed classes, then the \textit{Average Accuracy} is the average of all $Acc^t$ for all $T$ tasks:

\begin{equation}
Acc = \frac{1}{T} \sum_{t=1}^{T} Acc^t,
\end{equation}

Let \textit{Forgetting} ($\mathcal{F}^t$) be the difference between the model's highest accuracy on task $\mathcal{T}^t$ and its accuracy on task $\mathcal{T}^t$ at the end of all task training. Then, the \textit{Average Forgetting} is the average of $\mathcal{F}^t$ from task $\mathcal{T}^1$ to task $\mathcal{T}^{T-1}$:

\begin{equation}
\mathcal{F}^t = \max_{l \leq T} Acc^t_l - Acc^t_T,
\end{equation}

\begin{equation}
\mathcal{F} = \frac{1}{T-1} \sum_{t=1}^{T-1} \mathcal{F}^t,
\end{equation}

where $Acc^t_l$ represents the accuracy on task $\mathcal{T}^t$ at training step $l$ and $Acc^t_T$ represents the accuracy on task $\mathcal{T}^t$ after the training of all $T$ tasks.

\textbf{Training details.} To ensure a fair comparison, we adapted the federated learning algorithms \textbf{FedAvg}, \textbf{FedProx}, and the continual learning method \textbf{EWC} to the federated class incremental learning setting. Each client was trained with a batch size of 128 for a total of 100 communication rounds, with each round consisting of 5 local training epochs. We employed the SGD optimizer with a learning rate of 0.01, a momentum coefficient of 0.9, and a weight decay parameter of $5\times10^{-4}$. For other hyperparameters of the baseline methods, we adhered to the values recommended in their respective papers. For our method, we applied optimal hyperparameters derived through a grid-search strategy, setting $\alpha$ to 3 and $\gamma$ to 2. The diffusion models were trained for 200 epochs for each task using a batch size of 16, with the Adam optimizer and a learning rate of $5\times10^{-5}$. We generated images by denoising random noise over 1000 epochs by the sampling method DDPM. All results are averaged over three different random seeds.


\textbf{Environment} All our experiments were conducted on a single machine with 1TB RAM and 256-core AMD EPYC 7742 64-Core Processor @ 3.4GHz CPU. We use the NVIDIA H100 GPU with 80GB memory. The software environment settings are: Python 3.10.11, PyTorch 2.1.2 with CUDA 12.2 on Ubuntu 22.04.4 LTS.




\section{Code for Reproduction}

Our code is attached to the supplementary material.

\section{Analysis on Communication and Computation Cost}

\label{sec:cost_analysis}

\begin{table}[h]
    \centering
    \caption{Cost comparison on the CIFAR-100 dataset on $N$=10 clients when $T$=5.}
    \scalebox{0.9}{
\begin{tabular}{c|c|ccc}
\myhline
Metric & Communication Cost   & Client Training Time  & Server Runtime   &  Total Time \\
\hline
\textbf{FedAvg} &    \textbf{419.92 GB}   &  \textbf{3.13 h} &     \textbf{0.25 h}   &   \textbf{3.38 h}  \\
\textbf{FedProx} &   \textbf{419.92 GB}    &  4.19 h &     \textbf{0.25 h}   &  4.44 h  \\
\hline
\textbf{FedEWC} &   \textbf{419.92 GB}    &  4.46 h &     \textbf{0.25 h}   &  4.71 h  \\
\textbf{FLwF-2T} &   \textbf{419.92 GB}    &  3.28 h  &    \textbf{0.25 h}    &  3.53 h \\
\hline
\textbf{TARGET} &   463.87 GB    &  4.25 h   &    2.95 h    &   7.20 h  \\
\textbf{MFCL} &   581.05 GB    &  5.57 h &    1.24 h    &  6.81 h  \\
\textbf{\method\ (Ours)} &   \textbf{419.92 GB}    & 29.52 h &   \textbf{0.25 h}    &  29.77 h \\

\myhline
\end{tabular}%
}

    \label{tab:cost}
\end{table}%


Table \ref{tab:cost} presents the communication and computation costs for various methods based on ResNet-18 on the CIFAR-100 dataset. As we can see, our method exhibits significant advantages in terms of communication cost compared to other generative-based methods, which is a crucial concern in federated learning scenarios where minimizing communication overhead is essential. Specifically, the communication cost of our method is comparable to the baseline method FedAvg, maintaining a communication cost of 419.92 GB. This cost is considerably lower than that of TARGET and MFCL, thus enhancing the scalability of the federated learning system and improving data privacy by reducing the amount of information shared.


Additionally, as our method does not necessitate the server from training the generative models, the server runtime is notably short at 0.25 hours, comparable to that of FedAvg. This is particularly important when scaling a federated learning system to include a large number of clients (e.g., over 100 clients), as it alleviates server-side computational pressure. Although our method involves additional computational overhead and increases training duration on the client side due to the inherent limitations associated with the diffusion model, it offers more stable image generation and significant enhancements in model performance compared to GAN-based methods. Therefore, it is a trade-off between performance and computation cost. Moreover, the training and sampling times can be reduced by utilizing advanced techniques, such as DDIM, making the method more suitable for the FCIL scenarios.

\section{Details of the Diffusion Model}

\subsection{Model Structure}

In our method, we use the UNet~\citep{ronneberger2015u} as the base structure to train the diffusion models. UNet consists of an encoder-decoder structure with skip connections that facilitate the retention of high-resolution details during the image reconstruction process. The encoder systematically downsamples the input image while capturing spatial features, and the decoder subsequently upsamples the encoded representation, combining fine and coarse features via skip connections to reconstruct the image.

\subsection{Training Procedure}

We use the DDPM~\citep{ho2020denoising} method to train our diffusion models. The training of DDPM involves learning to denoise progressively noise-added images. This is formalized in the following steps:

\begin{enumerate}
    \item \textbf{Forward Diffusion Process:} We define a forward diffusion process to iteratively add Gaussian noise to a data sample $\mathbf{x}_0$ over $T$ time steps. At each time $t$, the noisy image $\mathbf{x}_t$ can be described as:
    \begin{equation}
        q(\mathbf{x}_t | \mathbf{x}_{t-1}) = \textbf{\textit{N}}(\mathbf{x}_t; \sqrt{\alpha_t} \mathbf{x}_{t-1}, (1-\alpha_t) \mathbf{I}),
    \end{equation}
    where $\alpha_t$ are predefined constants. 

    \item \textbf{Reverse Diffusion Process:} The objective of the DDPM is to learn the reverse process $p_{\theta}(\mathbf{x}_{t-1} | \mathbf{x}_t)$, which entails denoising the image $\mathbf{x}_t$ to recover $\mathbf{x}_{t-1}$. The reverse process is parameterized by a neural network (in our case, UNet), denoted by $\theta$. The likelihood of the training data is maximized by minimizing the following variational bound:
    \begin{equation}
        \mathcal{L} = \mathbb{E}_{q(\mathbf{x}_{1:T}|\mathbf{x}_0)} \left[ \sum_{t=1}^{T} \frac{1}{2\sigma_t^2} \left\| \mathbf{x}_t - \mathbf{\mu}_\theta(\mathbf{x}_t, t) \right\|^2 \right].
    \end{equation}
\end{enumerate}

\subsection{Sampling Procedure}
Once the model is trained, images can be generated via the learned reverse diffusion process. The sampling process reverts the sequence of noise addition steps from a pure Gaussian noise $\mathbf{x}_T$ back to the data space:

\begin{enumerate}
    \item \textbf{Initial Gaussian Sample:} Begin with $\mathbf{x}_T \sim \textbf{\textit{N}}({\bf 0},{\bf I})$.
    \item \textbf{Iterative Denoising:} For $t$ from $T$ to 1, use the trained model $p_\theta(\mathbf{x}_{t-1} | \mathbf{x}_t)$ to iteratively denoise $\mathbf{x}_t$ to recover $\mathbf{x}_{t-1}$. The denoising step can be expressed as:
    \begin{equation}
        \mathbf{x}_{t-1} = \frac{1}{\sqrt{\alpha_t}} \left( \mathbf{x}_t - \frac{1-\alpha_t}{\sqrt{1-\bar{\alpha}_t}} \mathbf{\epsilon}_\theta(\mathbf{x}_t, t) \right) + \sigma_t \mathbf{z},
    \end{equation}
    where $\mathbf{z} \sim \textbf{\textit{N}}({\bf 0},{\bf I})$ and $\mathbf{\epsilon}_\theta(\mathbf{x}_t, t)$ denotes the model prediction of the added noise.
\end{enumerate}

\section{Hyperparameter tuning for \method}

Hyperparameters are crucial in determining the performance of algorithms. Here, we demonstrate the sensitivity of algorithm performance to each hyperparameter. Our analysis involves modifying one parameter at a time while keeping the others constant. The hyperparameters we have tuned are:

\begin{itemize}
    \item $\lambda$: retention ratio for generated samples filtered by entropy.
    \item $\alpha$: weight of knowledge distillation loss (\(\mathcal{L}_{KD}\)).
    \item $\gamma$: weight of feature distance loss (\(\mathcal{L}_{FD}\)).
    \item $n_d$: number of epochs to train a diffusion model.
\end{itemize}

Table \ref{tab:hyperparams} presents the average accuracy \textit{Acc} and average forgetting $\mathcal{F}$ for each hyperparameter on the CIFAR-100 dataset with $T=5$ tasks. A minor discrepancy may exist between this value and the results presented in the main manuscript since the ablation was conducted using a single seed, while the results in the primary manuscript represent the average of three different seeds.

\begin{table}[h]
    \centering
    \caption{Effect of different hyperparameters for the CIFAR-100 dataset.}
    \label{tab:hyperparams}
     \scalebox{0.95}{
    \begin{tabular}{ccc|ccc|ccc|ccc}
        \myhline
        $\lambda$ & $\textit{Acc} (\uparrow)$ & $\mathcal{F} (\downarrow)$ & $\alpha$ & $\textit{Acc} (\uparrow)$ & $\mathcal{F} (\downarrow)$  & $\gamma$ & $\textit{Acc} (\uparrow)$ & $\mathcal{F} (\downarrow)$ & $n_d$ & $\textit{Acc} (\uparrow)$ & $\mathcal{F} (\downarrow)$  \\  
        \hline
        0.5  & 56.17 & 28.49 & 1 & 56.31 & 28.95 & 1 & 56.66 & 28.08 & 20 & 49.28 & 31.42 \\
        0.7  & 56.53 & 28.27 & 3 & 57.11 & 27.84  & 2 &  57.11 & 27.84  & 50 & 55.15 & 29.76 \\
        0.8  & 56.89 & 28.01 & 5 & 56.29 & 29.02 & 5 & 56.15 & 28.52 & 100 & 56.01 & 29.17 \\
        0.9  & 57.11 & 27.84 & 7 & 55.87 & 29.47 & 8 & 54.03 & 30.11 & 200 &  57.11 & 27.84  \\
        1.0  & 55.69 & 29.33 & 9 & 55.24 & 29.88 & 10 & 53.24 & 30.98 & 400 & 57.08 & 27.89 \\
        \myhline
    \end{tabular}
}
\end{table}

From Table \ref{tab:hyperparams}, it is evident that the performance of our method remains robust to variations in hyperparameters. Meanwhile, we have the following findings:

\begin{itemize}
    \item Setting the parameter $\lambda$ to a small value, such as 0.5, does not necessarily improve model performance. This is attributed to the fact that only samples with very low-confidence labels might adversely impact model training, whereas the majority of samples are generally suitable for effective training. 
    \item The values of $\alpha$ and $\gamma$ should not be excessively large, as this would harm the model performance. It is crucial to balance the cross-entropy loss with the knowledge distillation (KD) and feature distance (FD) losses. This balance enables the model to effectively learn from previous tasks without compromising its performance on the current task.
    \item Once the diffusion model has been sufficiently trained (i.e., $n_d \geq 200$), further training does not substantially improve the quality of the generated images or the model performance. This observation highlights the stability of the diffusion model as a generative tool.
\end{itemize}

The robustness of the model performance concerning each hyperparameter is crucial in both federated learning and continual learning, which remains a topic requiring further investigation. 

\end{document}